\documentclass{article}

\usepackage{PRIMEarxiv}

\usepackage[utf8]{inputenc} 
\usepackage[T1]{fontenc}    
\usepackage{hyperref}       
\usepackage{url}            
\usepackage{booktabs}       
\usepackage{amsfonts}       
\usepackage{nicefrac}       
\usepackage{microtype}      
\usepackage{lipsum}
\usepackage{fancyhdr}       
\usepackage{graphicx}       
\graphicspath{{media/}}     

\usepackage[square,numbers]{natbib}
\usepackage[title]{appendix}
\usepackage{comment}
\usepackage[noblocks]{authblk}

\usepackage{todonotes} 
\definecolor{editcolor}{cmyk}{0.0, 0.49, 1.0, 0.20}

\newcommand{\textoverline}[1]{$\overline{\mbox{#1}}$}

\title{Towards modeling evolving longitudinal health trajectories with a transformer-based deep learning model
\thanks{Under review} 
}

\author[1]{Hans~Moen}
\author[1]{Vishnu~Raj}
\author[2]{Andrius~Vabalas}
\author[3]{Markus~Perola}
\author[1,4]{Samuel~Kaski} 
\author[2,5,6]{Andrea~Ganna}
\author[1]{Pekka~Marttinen}

\affil[1]{Department of Computer Science, Aalto University, Finland}
\affil[2]{Institute for Molecular Medicine Finland, FIMM, University of Helsinki, Finland}
\affil[3]{Finnish Institute for Health and Welfare (THL), Finland}
\affil[4]{Department of Computer Science, University of Manchester, UK}
\affil[5]{Broad Institute of MIT and Harvard, Cambridge, MA, USA}
\affil[6]{Massachusetts General Hospital, Cambridge, MA, USA}
\affil[ ]{\texttt{hans.moen@aalto.fi, get.vichu@gmail.com, andrius.vabalas@gmail.com, markus.perola@thl.fi, samuel.kaski@aalto.fi, aganna@broadinstitute.org, pekka.marttinen@aalto.fi}}

\begin{document}
\maketitle

\begin{abstract}
Health registers contain rich information about individuals’ health histories. Here our interest lies in understanding how individuals' health trajectories evolve in a nationwide longitudinal dataset with coded features, such as clinical codes, procedures, and drug purchases. We introduce a straightforward approach for training a Transformer-based deep learning model in a way that lets us analyze how individuals' trajectories change over time. This is achieved by modifying the training objective and by applying a causal attention mask. We focus here on a general task of predicting the onset of a range of common diseases in a given future forecast interval. However, instead of providing a single prediction about diagnoses that could occur in this forecast interval, our approach enable the model to provide continuous predictions at every time point up until, and conditioned on, the time of the forecast period. We find that this model performs comparably to other models, including a bi-directional transformer model, in terms of basic prediction performance while at the same time offering promising trajectory modeling properties. We explore a couple of ways to use this model for analyzing health trajectories and aiding in early detection of events that forecast possible later disease onsets. We hypothesize that this method may be helpful in continuous monitoring of peoples' health trajectories and enabling interventions in ongoing health trajectories, as well as being useful in retrospective analyses.
\end{abstract}


\section{Introduction}
\label{sec:introduction}
Large electronic health registers, such as nationwide health registers, contain rich information about individuals' health histories.
Through the use of machine learning, we may be able to explore and better understand the health status and risk factors of individuals and groups in the population. This includes predicting future health outcomes.
However, the nature of such data introduces various challenges from a machine learning perspective \citep{si2021deep}.
In our case, the dataset includes many different age groups, it has large feature and label imbalances, and has missing information and ambiguity -- resulting from treatments, procedures, diagnoses, medications, and disease prevalence change as a result of the advancement of medicine.

Recent advances in deep learning have demonstrated great potential in handling intricate sequential data. A prime example of this lies in natural language processing. The advent of the transformer architecture \citep{vaswani2017attention} has had a big impact on the state-of-the-art, as well as enabled new applications. 
Transformer-based models have also been applied to longitudinal and time-series health record data (see, e.g., \citep{rasmy2021med, munoz2023interpreting}). However, in disease onset prediction tasks, transformer-based models do not necessarily seem to outperform simpler models \citep{li2022generic, lam2022multitask, antikainen2023transformers}.

Here, we focus on a nationwide dataset containing longitudinal health data with coded features from various registers, such as clinical codes, procedures, and drug purchases (see Section~\ref{sec:data})
The task we focus on is about predicting, for individuals, the onset of common diagnoses during a five-year window based on their previous health history. A person's health history consists of a discrete time series of codes. We refer to the interval used as the basis for prediction (i.e., the input to the model) as the \emph{historical interval}, and the follow-up period, in which we predict the onset of diagnoses and negative life events, is referred to as the \emph{forecast interval}.

\begin{figure*}[htb]
    \centering
    \includegraphics[width=0.9\linewidth]{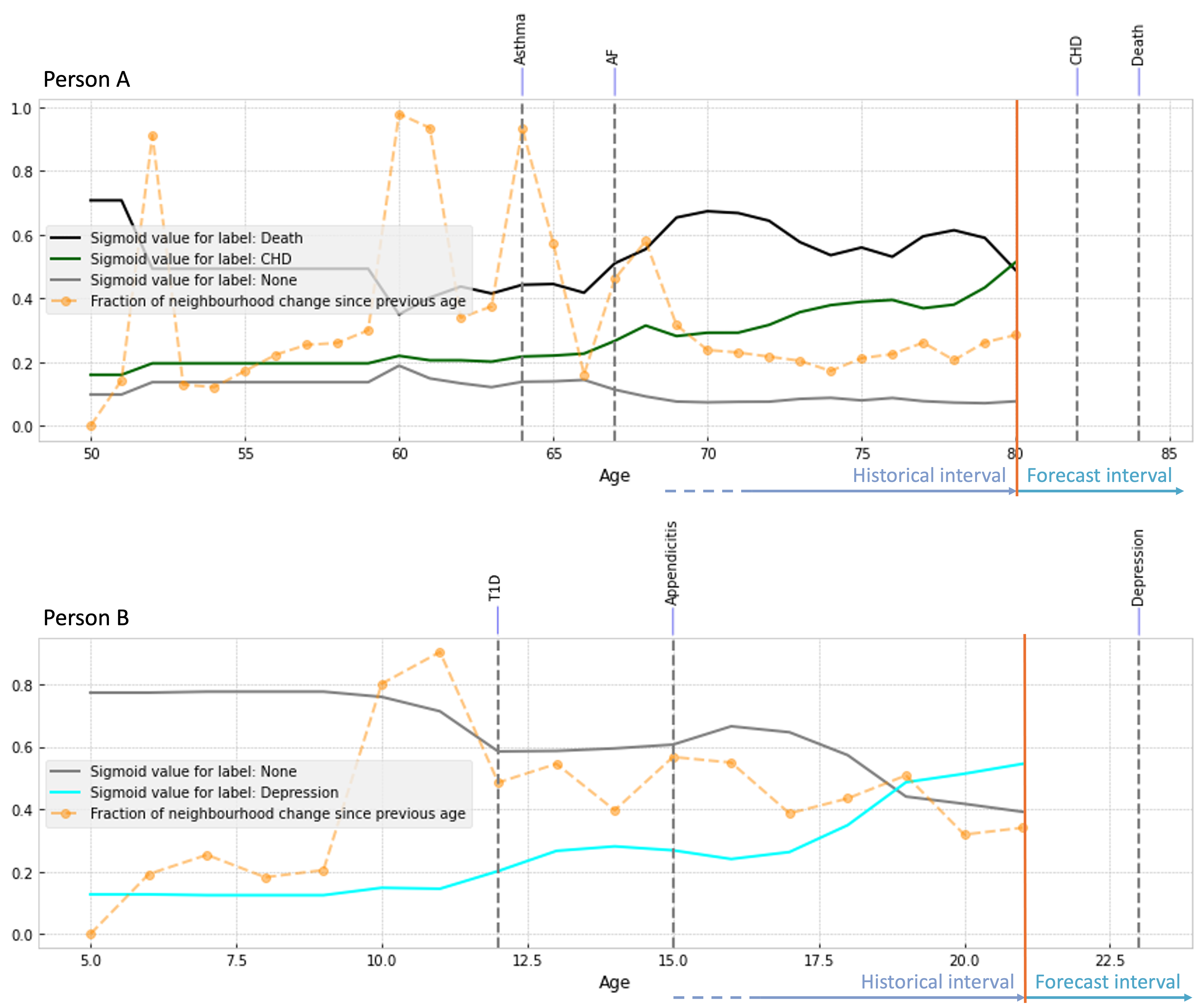}
    \caption{Figure shows an example of two artificial persons' health progressions. The plots shows how the model's prediction probabilities -- sigmoid values -- for the various class labels (diseases to occur in the forecast interval) changes over time as they age. Here we have also included the sigmoids for the \emph{none} class. The orange dots indicate the fraction of changes in the nearest neighborhood from one age to the next, calculated from age-wise embedding similarities to the $k$-nearest individuals.}
    \label{fig:sigmoid_and_nn_changes}
\end{figure*}

In this work, we introduce a transformer-based approach for modeling discrete-time health trajectories with an inherent property that enables modeling and analysis of health trajectories as they change over time.
Instead of providing a single outcome prediction, it makes an updated prediction for every code in the time series of an individual relative to the given future forecast interval.
We show how the model can provide insights into early predictors of later onset of diseases and negative life events.
Figure~\ref{fig:sigmoid_and_nn_changes} shows an example of how the model's prediction probabilities -- sigmoid values -- for the various class labels (diagnoses to occur in the forecast interval) changes over time as they age. Here we have also included the sigmoids for the \emph{none} class.
Further, the orange plots indicate changes in the neighborhood from one age to the next according to similarities calculated from the latent representations of the model.

In addition to reporting basic prediction performance, we explore a few ways of utilizing the model to detect when and why a change in a person's health trajectory occurs. This is done by either: a) analyzing changes in the model's prediction probabilities during the forecast interval -- from one year to the next; or b) analyzing how similarities between individuals change over time. 
A summary of contributions in this paper:
\begin{itemize}
    \item We present a transformer-based model variant tailored for modeling evolving health trajectories from a nationwide longitudinal dataset. 
    \item We propose several ways of using this model to analyze health trajectories and to aid in early detection of events that signal changes in individuals' health status and subsequent disease onsets.
\end{itemize}


\section{Related Work}
\label{sec:related_work}
Predicting future health risks using Electronic Health Records (EHRs) data is a well-established area of research. Models like Logistic Regression and Support Vector Machines (SVMs) have been applied \citep{ali2019optimized, chu2020endpoint}. The same goes for tree-based methods like eXtreme Gradient Boosting (XGBoost) \citep{ogunleye2019xgboost, raihan2023detection, yi2023xgboost, wu2022interpretable, zang2023early}. However, these may not optimally capture complex temporal relationships in longitudinal EHR data. Recurrent Neural Networks (RNNs), such as Long Short-Term Memory (LSTM) and Gated Recursive Unit (GRU) networks, can model sequential data \citep{guo2021predicting, vabalas2023}, but may struggle with long-range dependencies \citep{xie2022deep}.



Attention-based transformer models represent the latest generation of sequential models, and have also been used in healthcare data analysis. Transformers are popular for their scalable parallel training capabilities and ability to capture complex dependencies within sequences, making them well-suited for EHR data. 
Studies have shown that transformer-based models perform well in tasks like predicting hospital readmission \citep{miao2023must,sheetrit2023predicting} and identifying patient phenotypes \citep{wang2024patient}. 
Notably, the BEHRT model leverages transformers to effectively model sequential EHR data for tasks like disease pattern retrieval and postoperative mortality prediction \citep{li2020behrt}. 
\citet{li2022hi} proposed Hi-BEHRT model which improves disease risk prediction by using hierarchical transformers to handle long EHR sequences.
Still, transformer-based models do not necessarily outperform simpler models in basic health outcome prediction tasks \citep{li2022generic, lam2022multitask, antikainen2023transformers}.
Other works have explored variants of the transformer architecture \citep{choi2020learning, kumar2024self}.

Several works focus on generating and analyzing static (global) code embeddings from EHR data \citep{hong2021clinical, finch2021exploiting}.
%
%
%
\citet{hettige2019mathtt} introduced MedGraph, a graph-based model that captures both the relationships between codes within a visit and the sequencing of visits over time. 
%
\citet{lee2020temporal} introduced an RNN-based approach for clustering time-series data that encourages latent representations (embeddings) of patients with similar future outcomes of interest to have similar representations at each time step.
%
\citet{munoz2023interpreting} explores ways of interpreting the embedding space of a bidirectional transformer-based model trained on EHR data. By projecting the embeddings down to two dimensions, they show that the model generates similar representations for similar comorbid-diseases or medications.
They also explore temporality by segmenting the data into fixed time windows (snapshots of 5 or 10 years) around when a target disease was assigned, before analyzing changes between the embeddings from the different time windows. They show how this can be used to identify subgroups within a disease who have distinct progression patterns.
\citet{chen2023interpreting} proposes a contrastive explainability approach (XAI) that focuses on identifying the most influential features affecting the latent state of the model. They show how this can be applied to a state space model to help discover features that are influential to the latent space over time.

The present work may be seen as an indirect approach to change-point detection (see, e.g., \citep{dion2023multiple, gupta2022real}). However, the main difference between our approach and the common change-point detection approaches is that we focus on modeling continuously evolving changes in the predictions, instead of focusing on discontinuities in the data distribution.


Current approaches for disease prediction based on transformer models typically use models that are bidirectional in nature, and thus expect the entire or a fixed time window of the historical event interval as input for predicting future disease onsets.
This limits their ability to capture and interpret the evolving nature of health trajectories. 
We propose a simple approach that attempts to address this. We are not aware of others who have trained a non-generative transformer model with a causal attention mask the way we do. This allows us to analyze each time step and how the information up to each time step may influence predictions of future events. We are neither aware of previous work that analyzes temporal changes in the model's predictions or in the latent embedding neighborhood in the way we explore here.

\section{Methods}
\label{sec:methods}

\subsection{Unidirectional Transformer Model with Evolving Predictions}
\label{subsec:models}
The base machine learning architecture used is the transformer with multiple stacked multi-headed self-attention layers \citep{vaswani2017attention}.
Each element in an input sequence is computed in parallel, which differs from recursive models.
The originally proposed transformer model/architecture combined both an encoder and decoder component into an encoder-decoder model. 
Afterwards, we have seen models utilizing only encoders, such as BERT \citep{devlin2019bert}, with bidirectional attention, which are typically used for classification tasks.
Decoder-only models have become popular as generative models, such as GPT \citep{radford2018improving,radford2019language}. These incorporates causal attention masking to make the model unidirectional with generative capabilities.

\begin{figure}[htb]
      \centering
      \includegraphics[width=0.7\linewidth]{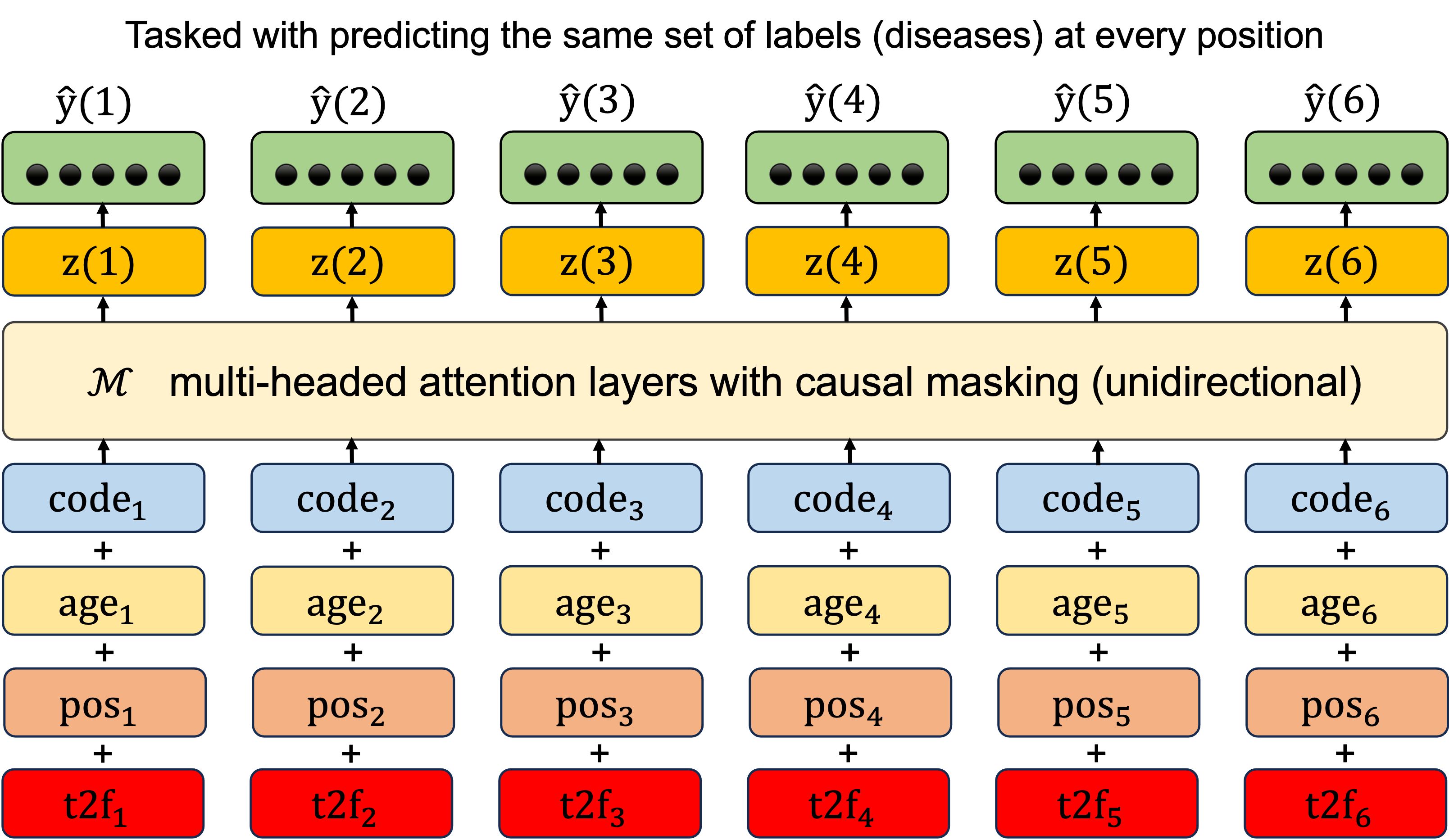}
      \caption{The figure illustrates the model and its inputs: codes, ages, positional information (pos), and years to forecast (t2f). A decision layer with sigmoid activations enables the model to predict the targeted labels -- reflecting what will happen in the forecast interval -- at each input position.}
      \label{fig:model}
\end{figure}

Inspired by previous work on transformer models for EHR data, the inputs to the models are the codes ($\textbf{code}$), the associated ages of the person when these codes occur ($\textbf{age}$), and their position in the sequence ($\textbf{pos}$, see e.g. \citet{li2020behrt}). We also introduce an additional parameter which represents the number of years until the forecast interval ($\textbf{t2f}$). 
Codes, ages, positional information, and years to the forecast period go through an embedding look-up layer before these are position-wise summed and given as input to the model, as illustrated in Figure~\ref{fig:model}.
The input to the model for one individual $n$, with $T_{n}$ codes in their historical interval, is thus\\ $\textbf{x}_{n}(1:T_{n}) = (\textbf{code}_{1:T_{n}}, \textbf{age}_{1:T_{n}}, \textbf{pos}_{1:T_{n}}, \textbf{t2f}_{1:T_{n}})$.
Each individual has one to multiple labels ($\textbf{y}_{n}$) indicating possible diseases, \emph{death}, or \emph{none}, occurring in the forecast interval. Since we have a fixed set of classes $C$, the labels for one individual are represented as a binary vector $\textbf{y}_{n} \in [0,1]^{C}$.

\paragraph{The \texttt{Evolve} model}
The motivation for this approach is to enable the model to identify and predict, as early as possible in a person's life, upcoming health problems. This ultimately allows us to study how these predictions change (evolve) over time.
To achieve the desired properties, we apply an attention mask as in decoder-based transformer models, \textbf{however we do not train it in a next-token prediction manner}.
Instead, during training, the model is tasked with always predicting, for every input $\textbf{x}_{n}(1:t)$ with $t \in 1 \ldots T_{n}$, the same set of diagnosis labels $\textbf{y}_{n}$ that occur in this person's forecast interval.
This means that we also have $T_{n}$ predictions, $\hat{\textbf{y}}_{n}(1:T_{n})$.
Attention masking makes the prediction $\hat{\textbf{y}}_{n}(t)$ conditioned only on $\textbf{x}_{n}(1:t)$. During training, $\textbf{y}_{n}(t)$ is the same for all $t \in 1 \ldots T_{n}$.
For individual $n$, the model $\mathcal{M}$ will thus make $T_{n}$ multi-label predictions:
\begin{equation}
    \mathcal{M}(\textbf{x}_{n}(1:T_{n})) = \hat{\textbf{y}}_{n}(1:T_{n})
\end{equation}

We apply a decision layer (decoder) with sigmoid activation on top of the output embeddings from the final layer at each position $t$, which maps these to the probabilities for the labels $\hat{\textbf{y}}_{n}(t)$.
Since the prediction task is multi-label classification, the loss function used is binary cross-entropy, $\mathcal{L}_{\textrm{BCE}}$, which is applied at every position $t$:
\begin{equation}
    \mathcal{L}_{n}(t) = \frac{1}{C}\sum_{c=1}^{C}{\mathcal{L}_{\textrm{BCE}}(\textbf{y}_{n}^{c}(t), \hat{\textbf{y}}_{n}^{c}(t))}
\end{equation}
Where $\hat{\textbf{y}}_{n}(t) = \mathcal{M}(\textbf{x}_{n}(1:t))$. And the total loss for one person becomes:
\begin{equation}
    \mathcal{L}_{n} = \frac{1}{T_{n}}\sum_{t=1}^{T_{n}}{\mathcal{L}_{n}(t)}
\end{equation}

\subsection{Reference Models}
\label{subsec:baselines}
Primarily to verify that the masking and training approach used in the \texttt{Evolve} model does not have a large negative impact on its overall prediction performance, we also include a couple of reference models. 
One model is a bidirectional transformer-based BERT-style model \citep{devlin2019bert, li2020behrt} that we simply call \texttt{CLS}. This does not have attention masking and has a special "CLS" token as the first code position in the sequence. A decision layer is (only) applied to the output from this CLS token when predicting labels and calculating loss. 
We also include a Logistic Regression model \citep{hosmer2013linreg} (\texttt{LogReg}) and a XGBoost model \citep{chen2016xgboost} (\texttt{XGBoost}), which are trained on count-based feature (code) vectors concatenated with age.

\subsection{Extracting Age Embeddings}
Transformer models are known for their contextualized embedding representations resulting from the attention mechanism.
One way of analyzing health trajectories involves looking at the latent embedding space of the \texttt{Evolve} model.
To do so, we first aggregated the output embeddings at each position into age embeddings ($a \in \{A_{n}\}$).
This is done by applying Position-Weighted Mean Pooling ($\textrm{PWMPooling}$) to the set of embeddings at each age ($\textbf{z}_{n}$).
This is similar to what \citet{muennighoff2022sgpt} used to extract semantic sentence embeddings from a GPT model.
\begin{equation}
    \textbf{z}_{n}(a) = \textrm{PWMPooling}(\{\textbf{z}_{n}(t):\textbf{age}_{t}=a\})
\end{equation}
Weights are $1, 2, 3, \ldots m$, where $m$ is the number of inputs at age $a$. Most recent embedding gets the largest weight. The embeddings are finally normalized.
As individuals may not have recorded data at every age, we simply fill in the gaps with the embeddings from the preceding ages for the subsequent analysis.

\subsection{Computing Nearness and Neighborhood Changes in the Embedding Space}
\label{subsec:neighbourhood_changes}
To compute the nearness between pairs of age embeddings we used the \emph{cosine similarity metric} ($cos$), which quantifies the similarity between two vectors (embeddings) to a value in the range 0--1, low--high. 
One application of this is to calculate the similarity between two individuals, for example $n$ and $m$, at ages $a$ and $b$; $cos(\textbf{z}_{n}(a), \textbf{z}_{m}(b))$.

We also use this to compute the rate of change in the neighborhood of a target individual as they age. More specifically, we iteratively calculate the fraction of changes among reference individuals that are most similar to the target individual, from one age ($a-1$) to the next ($a$). $\mathcal{N}_{n}(a, k)$ which is the $k$ nearest neighbors to $n$ (individuals with the highest cosine similarity) when all are at age $a$. The rate of change at age $a$ with neighborhood size $k$ is thus:
\begin{equation}
    \label{eqn:rate_of_change}
    r_{n}(a, k) = 1 - \frac{|\mathcal{N}_{n}(a-1, k) \cap \mathcal{N}_{n}(a, k)|}{k}
\end{equation}
As an example, the orange dots in Figure~\ref{fig:sigmoid_and_nn_changes} shows the rate of change from one year to the next in these persons' health histories.

\section{Data}
\label{sec:data}
The dataset used is a collection of Finnish nationwide registers -- FinRegistry \citep{viippola2023data}. FinRegistry is a comprehensive register-based data resource that provides access to a diverse range of health and sociodemographic data for the entire Finnish population.
The initial dataset included 7,166,416 individuals, with 74.51\% alive on January 1, 2010.
We excluded individuals who died before the forecast interval, emigrated, or had no healthcare records leaving a sample of 5,173,795: for training (70\% - 3,620,947) validation (10\% - 517,160) and testing (20\% - 1,035,688).
At the beginning of 2016 (the start of the forecast interval), the mean age of our study population was 41.9 years (SD=24.3). There were slightly more females (50.8\%) than males (49.2\%).
For training we have used the data up to 2015.09.01 (historical interval), and the five-years period 2016.01.01--2020.12.31 was used for predictions (forecast interval). We left a three-month buffer between the two periods to avoid potential outcome information leakage.

In the forecast interval (2016--2020), we are predicting 21 possible outcomes (classes); 19 diagnoses, \emph{death}, and \emph{none}. 
Each individual will have a fixed set of labels $y_{n}$ for the given forecast interval. 
All diagnosis labels indicate the first time that a person is given the diagnosis. Thus, if a person has the label \emph{none}, it means that no new diagnoses occurred in the forecast interval. The 19 diagnoses are selected from what is termed \emph{clinical endpoints}. These are highly curated clinical disease definitions, predominantly generated by combining ICD[8-10] records coming from Healthcare, Causes of death, and Cancer registers. 
The selection of these diagnoses was deliberate, aiming to encompass a substantial portion of the disease burden in the population and exerting a significant impact on Disability Adjusted Life Years (DALY). 

\begin{figure}[htb]
      \centering
      \includegraphics[width=0.5\linewidth]{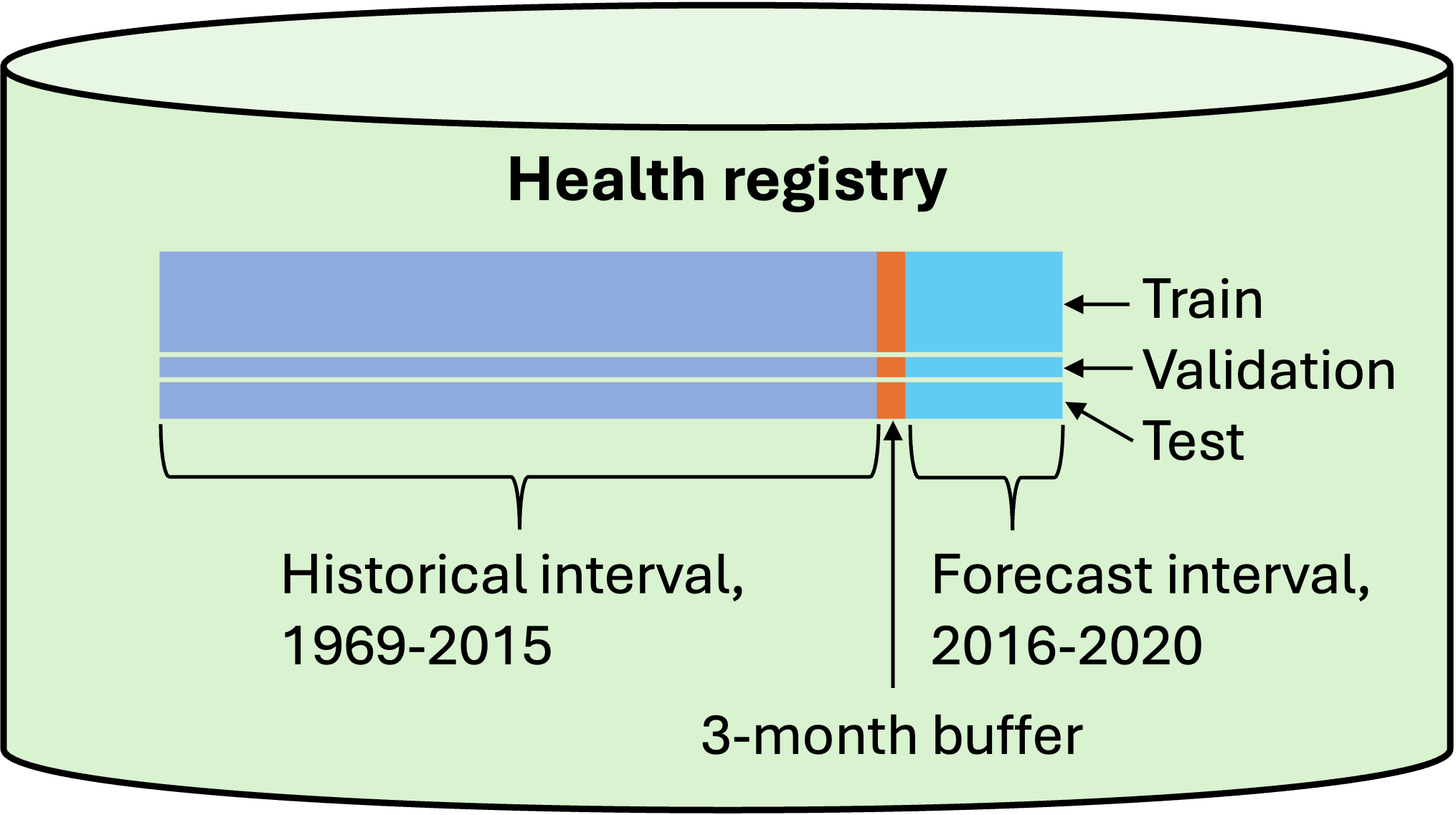}
      \caption{The dataset was split into historical and forecast intervals and further split into train, validation, and test sets.}
      \label{fig:data}
\end{figure}

Figure~\ref{fig:data} illustrates how the dataset was split into historical and forecast intervals, which was further split into train, validation, and test sets.

Table~\ref{tab:code_types} shows the different code types and their counts.
\begin{table*}[htb]
    \small
    \caption{Code types and their counts}
    \centering
    \begin{tabular}{ll}
        \toprule
        \textbf{Code sources (types)} & \textbf{Uniques} \\
        \midrule
        Primary care causes for visits  & 1092 \\
        Drug purchases (ATC code)  &  440 \\
        Clinical endpoints  & 2864 \\
        ICD - primary care ICD-10 codes  & 1529 \\
        Infectious disease (grouped by bacteria causing the disease) &   87 \\
        Nomesco - secondary care surgical procedures  & 1697 \\
        Primary care procedures  &  361 \\
        \bottomrule
    \end{tabular}
    \label{tab:code_types}
\end{table*}

See Table~\ref{tab:prevalence}, Appendix~\ref{apd:second}, for more detailed information on the classes (labels) and their prevalence.

\section{Experiments and Results} 
\label{sec:experiments}
Here we present a set of experiments and their results.

\subsection{Model Training and Performance}
\label{subsec:exp_model_training_and_performance}
For implementation of the transformer-based models (\texttt{Evolve} and \texttt{CLS}), our starting point was the nanoGPT implementation by \citet{nanogpt}. Other Python libraries and packages used for implementation and evaluation include Pytorch~\citep{paszke2019pytorch}, NumPy~\citep{harris2020array}, Scikit-learn~\citep{pedregosa2011scikit}. 
The \texttt{XGBoost} model was trained using the xgboost Python library in combination with Scikit-learn. For the \texttt{LinReg} model we used the implementation available through Scikit-learn. The MultiOutputClassifier wrapper was used to enable multi-label classification for the latter two models. 
See Appendix~\ref{apd:model_training} for more information on model training.

The results of predicting the labels in the forecast interval, given the entire historical interval ($1 \ldots T$) for each individual in the test set as input, are shown in Table~\ref{tab:performance}.
Evaluation metrics employed were area under the receiver operating characteristic curve (AUROC), area under the precision-recall curve (AUPRC), and Recall@4. The AUROC metric provides an overview of the model's ability to discriminate between positive and negative instances across varying decision thresholds, with higher AUROC values indicating superior discriminatory power. 
The AUPRC metric offers insight into the precision-recall trade-off, measuring the weighted average of precision achieved at each recall level. 
Recall@4 focuses on capturing the model's ability to correctly identify positive instances within the top four predictions according to sigmoid values, which is particularly relevant in scenarios where prioritizing high-recall predictions among the top-ranked results is crucial. 
Further, for each of these we report both \emph{micro}-averaging (equal weight to each instance) and \emph{macro}-averaging (equal weight to each class). Given the imbalanced nature of the data and that the task is multi-label classification, we believe the macro scores are more informative than micro scores.
Finally, Table~\ref{tab:prevalence} in Appendix~\ref{apd:second} shows the prevalence and AUROC of each individual class, calculated using the \texttt{Evolve} model.

\begin{table*}[htb]
    \centering
    \small
    \caption{Performances of the various models on the test set. We used bootstrapping to calculate standard deviations (sampling with replacement, 1k iterations), shown as subscripts.}%
    \begin{tabular}{l|ll|ll|ll}
        \toprule
        \textbf{Model}& \multicolumn{2}{c}{\textbf{\texttt{AUROC}}} & \multicolumn{2}{c}{\textbf{\texttt{AUPRC}}} & \multicolumn{2}{c}{\textbf{\texttt{Recall@4}}} \\
        & micro & macro  &  micro & macro  &  micro & macro\\
        \midrule
        \textbf{\texttt{XGBoost}} & 0.9799$_{0.0001}$ & 0.8413$_{0.0007}$  &  0.8475$_{0.0004}$ & 0.1291$_{0.0005}$  &  0.9368$_{0.0003}$ & 0.4262$_{0.0013}$ \\

        \textbf{\texttt{LogReg}} & 0.9781$_{0.0001}$ & 0.8305$_{0.0007}$  &  0.8308$_{0.0005}$ & 0.1213$_{0.0005}$  &  0.9323$_{0.0003}$ & 0.4033$_{0.0011}$ \\

        \midrule
        
        \textbf{\texttt{CLS}} & 0.9792$_{0.0001}$ & 0.8357$_{0.0007}$  &  0.8433$_{0.0004}$ & 0.1227$_{0.0004}$  &  0.9353$_{0.0003}$ & 0.4373$_{0.0013}$ \\
        
        \textbf{\texttt{Evolve}} & 0.9787$_{0.0001}$ & 0.8354$_{0.0007}$  &  0.8378$_{0.0005}$ & 0.1228$_{0.0004}$  &  0.9358$_{0.0003}$ & 0.4353$_{0.0013}$ \\

        \bottomrule
    \end{tabular}
    \label{tab:performance}%
\end{table*}

\subsection{Detecting Trajectory Changing Events} 
\label{subsec:exp_event_detection}
Here we propose two ways to use the \texttt{Evolve} model to detect important events and codes in individual's health histories that are associated with changes in their health status.
%
\paragraph{Looking at the changes in the model's prediction probabilities (sigmoid values)}
This implies analyzing how the sigmoid values for each label evolve over time, as the model iteratively sees more of an individual's health history. This provides insight into important codes and code sequences that trigger any changes to these sigmoid values.
In Figure~\ref{fig:sigmoid_and_nn_changes} the plots show examples of how the predicted probability for various diseases changes over time for this person relative to the forecast interval.
Important events (input codes) associated with a label may be found by sampling codes causing increases in the sigmoid value, or when the sigmoid value passes a predefined decision threshold. In this experiment we explore the former approach.

First, we presented the health history $\textbf{x}_{n}(1:T_{n})$ of each individual $n$ in the validation set to the model.
From these we extracted the maximum jump in sigmoid value (from one code to the next) for each class where $y_{n}^{c} = 1$. From this set, we accumulated the mean jump per code $\bar{j}_{1:C}$.
Next, for each individual in the test set, we sampled the sigmoid jumps per class $c$ that were equal to or greater than $\bar{j}_{c}$.
In Table~\ref{tab:early_examples} we show the top diagnoses (endpoints) and drug codes that are found to correlate with jumps for the various classes. We also include the average age of when the codes were given and the average number of years to the forecast interval (t2f). See Table~\ref{tab:early_examples_cont} in the Appendix for additional examples.


\begin{table*}[htb]
    \centering
    \scriptsize 
    \caption{Most frequent diagnosis codes and drug codes that triggers jumps in the sigmoid values of the respective classes (labels). Percentages are relative to total detected jumps per class.}%
    \begin{tabular}{l|llll|llll}
        \toprule
        \textbf{Class} & \textbf{Diagnosis codes} & \textbf{\%} & \textoverline{\textbf{age}} & \textoverline{\textbf{t2f}} & \textbf{Drug codes} & \textbf{\%} & \textoverline{\textbf{age}} & \textoverline{\textbf{t2f}}  \\
        \midrule
        & Type 2 diabetes & 6.12 & 61 & 10                              & Beta blocking agents, selective & 5.46 & 78 & 5 \\
        CHD & Heart failure, not strict & 3.05 & 76 & 7                 & Organic nitrates & 3.46 & 80 & 7 \\
        & Atherosclerosis & 2.24 & 68 & 7                               & Sulfonamides, plain & 3.18 & 80 & 4 \\
        \midrule
        & Substance abuse & 7.64 & 46 & 8                               & Anticholinesterases & 1.54 & 69 & 3 \\
        Epilepsy & Any mental disorder & 3.41 & 28 & 5                   & Sel. serotonin reuptake inhibitors & 1.11 & 59 & 5 \\
        & Alzheimer’s disease & 3.02 & 70 & 3                           & Other antipsychotics & 1.02 & 46 & 4 \\
        \midrule
        & Emotional disorders specific to childhood & 2.74 & 13 & 5     & Sel. serotonin reuptake inhibitors & 25.99 & 30 & 7 \\
        Depression & Any mental disorder & 2.25 & 22 & 2                & Other antidepressants & 11.78 & 35 & 4 \\
        & Anxiety disorders & 1.91 & 25 & 5                             & Benzodiazepine related drugs & 1.32 & 29 & 5 \\
        \bottomrule
    \end{tabular}
    \label{tab:early_examples}
\end{table*}

\paragraph{Looking at when there are high rates of change in the latent embedding neighborhood}
Observing a shift in the neighborhood of a person, in the model's embedding space, may indicate that a person's overall health status has changed relative to the reference group. 
We hypothesize that important events in a person's life may result in noteworthy changes in the affected individual's neighborhoods.
As a way to explore this, we have first identified a group of mothers who lost their child (target group), as well as an associated control group with mothers who did not lose their child.
We used propensity score matching to identify similar aged mothers with the kids born at the same age\footnote{Matching criteria included age, sex of child, age of parents at child birth, and number of children in the family.}.

\begin{figure*}[!htb]
      \centering
      \includegraphics[width=0.95\linewidth]{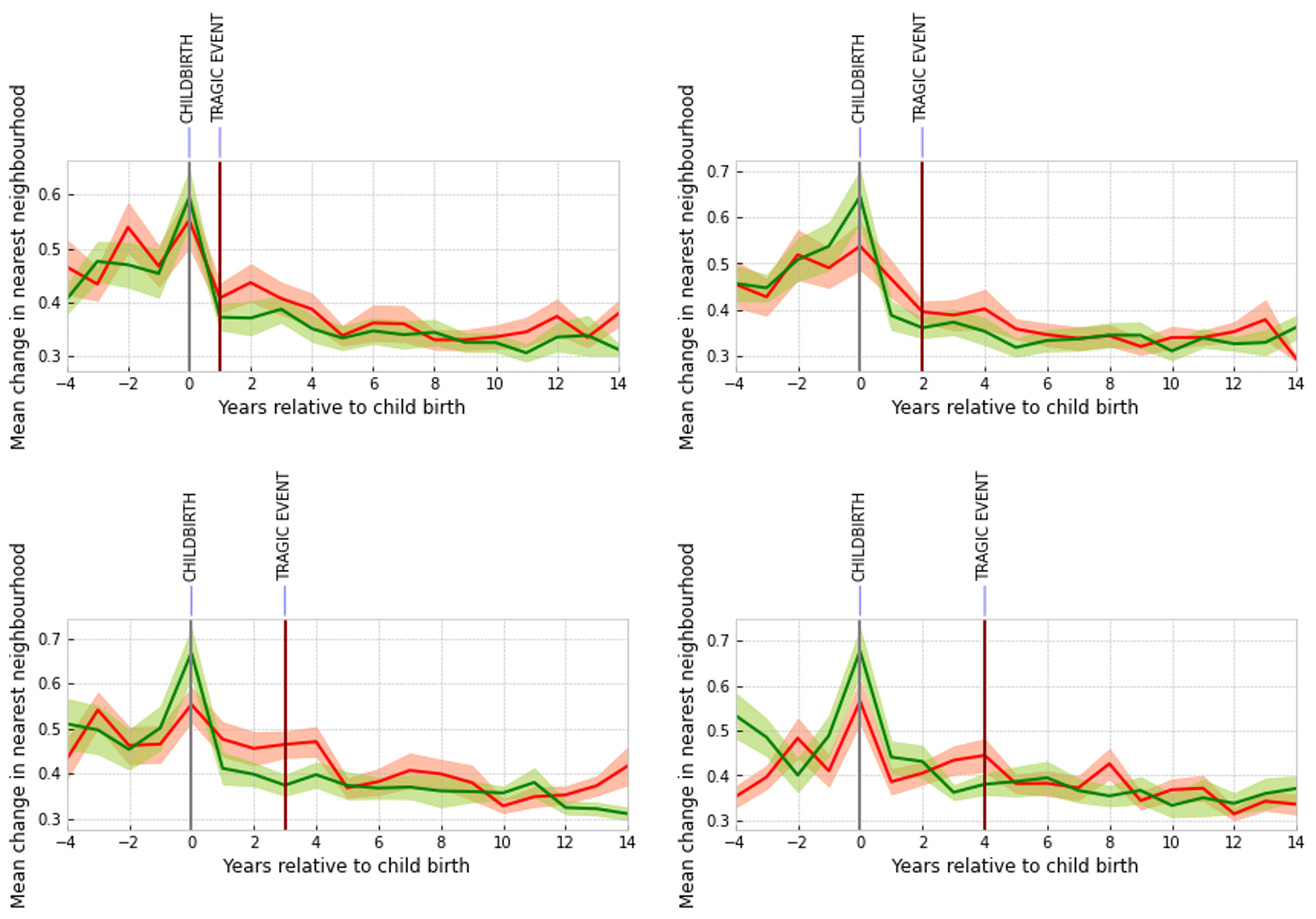}
      \caption{Average neighborhood change from one year to the next. The included individuals are mothers who did (red, target group TG), and did not (green, control group CG), lose their child at a specific age. Death at age 1, $n_{\mathrm{TG}}=27$ \& $n_{\mathrm{CG}}=27$ (upper left); death at the age 2, $n_{\mathrm{TG}}=18$ \& $n_{\mathrm{CG}}=18$ (upper right); death at age 3, $n_{\mathrm{TG}}=23$ \& $n_{\mathrm{CG}}=23$ (lower left); death at age 4, $n_{\mathrm{TG}}=19$ \& $n_{\mathrm{CG}}=19$ (lower right). Neighborhood size $k$ is set to 1000.}
      \label{fig:tragic_events}
\end{figure*}

Figure~\ref{fig:tragic_events} shows the mean change in the embedding neighborhood per group, from one year to the next, around the events of child birth and child death. Change is calculated as in eq.~\ref{eqn:rate_of_change}, and we use a neighborhood size of $k=1000$. 

\subsection{Nearest Neighbors Analysis}
\label{subsec:exp_nearest_neighbours_analysis}
Here we propose another way of exploring and interpreting individual's health trajectories with the \texttt{Evolve} model.
This too involves looking at their nearest neighbors in the latent embedding space as they age.
However, the aim here is to analyze \textbf{how a person's health profile becomes more or less similar to representatives from each disease class over time}.

\begin{figure*}[!htb]
      \centering
      \includegraphics[width=0.9\linewidth]{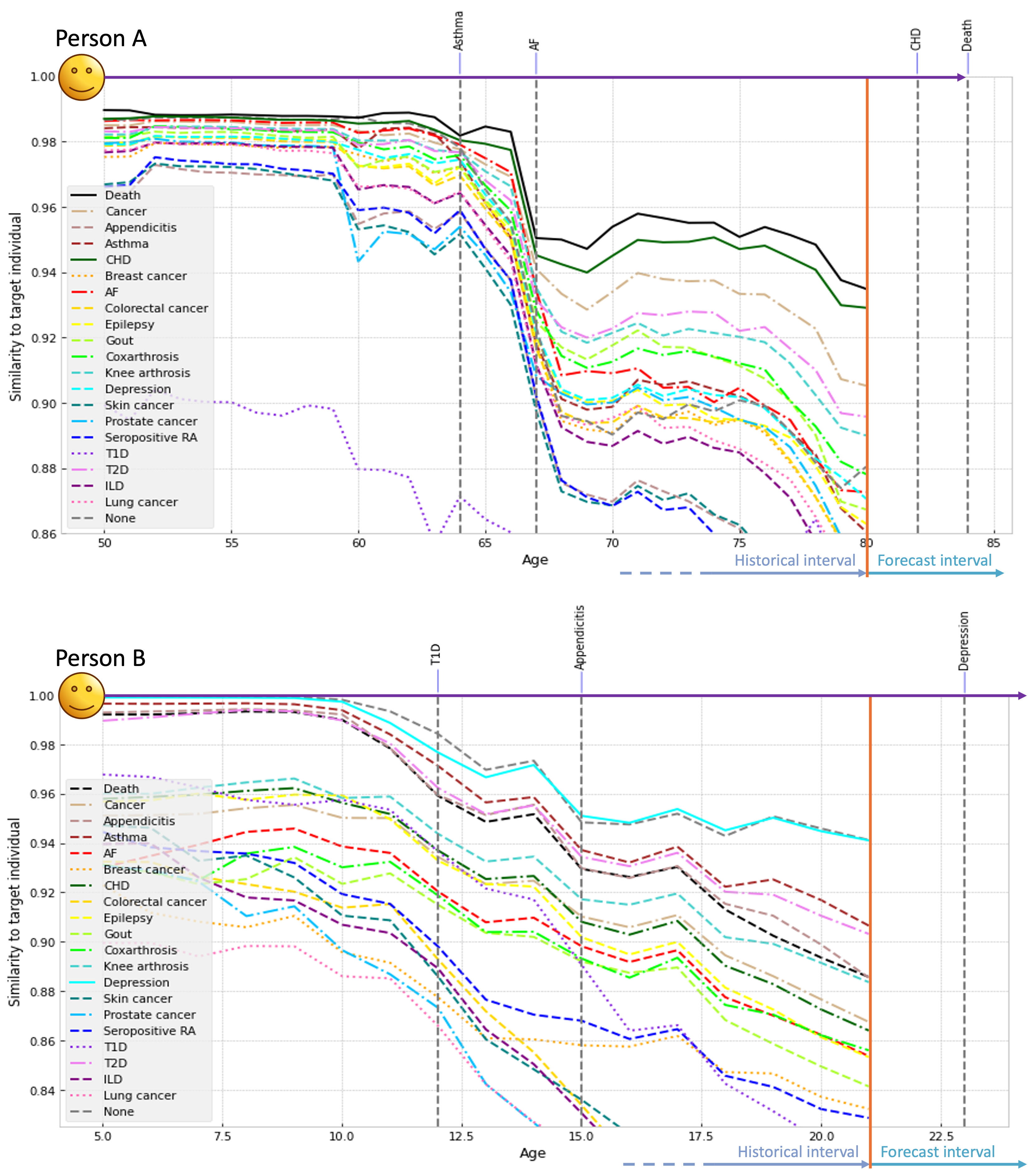}
      \caption{The figure shows the evolving age-wise similarities between the target individuals and the top $k=100$ most similar representatives from each diagnosis (class), calculated using the associated embedding representations from the \texttt{Evolve} model. This shows an alternative way of visualizing the health trajectories of the same artificial persons as in Figure~\ref{fig:sigmoid_and_nn_changes} (Person A and Person B).}
      \label{fig:nearness_trajectory_ageWise}
\end{figure*}

We observed that all individuals, even with the same outcome labels, tend to have quite unique health trajectories, especially as they get older. This increasing uniqueness of individuals' health trajectories is also observed in \citet{jylhava2017biological}.
Thus, instead of pre-calculating global diagnosis centroids for each diagnosis, we found that it made more sense to only compare the target person to a limited set of the \emph{most similar representatives} from each diagnosis.
For a single target person, at every age, we calculate how similar his or her embedding profile is to the mean embedding of the top $k$ most similar reference individuals from each class ($\{y^{1}\}, \{y^{2}\} \dots \{y^{C}\}$, i.e., we know they will get the diagnosis $y^{c}$) when they were the same ages\footnote{One reference individual may belong to multiple reference groups if they have more than one label.}.
Figure~\ref{fig:nearness_trajectory_ageWise} shows examples of how the similarities between target persons and references from each class evolve over time, left to right.
These are the same individuals as seen in Figure~\ref{fig:sigmoid_and_nn_changes}.
In the beginning, the target persons' profiles are similar to most other reference individuals, but over time, in particular around when getting some of the annotated diagnoses during the historical interval, they start becoming more or most similar to individuals who will have a similar health outcome in the forecast interval -- hearth problems and death for Person A (top), and depression for Person B (bottom).


\section{Discussion} 
\label{sec:discussion}
From the performance evaluation in Section~\ref{subsec:exp_model_training_and_performance}, we can conclude that the \texttt{Evolve} model performs comparable to the \texttt{CLS} model/approach in terms of predicting disease onsets in the forecast interval given the full historical interval as input. This means that the applied masking and training approach does not appear to have a notable negative impact on the overall predictive performance of the model. 
We also observe that these models perform comparable to, and slightly worse than, the \texttt{XGBoost} model. This is similar to what was observed in \citet{li2022generic, lam2022multitask, antikainen2023transformers}.


From the experiment on detecting important predictive health events, Section~\ref{subsec:exp_event_detection}, we observe that the identified codes in Table~\ref{tab:early_examples}, and in Table~\ref{tab:early_examples_cont}, seem to be sensible predictors of the associated labels.
This seems promising and warrants further exploration of the model and the underlying medicine.
Here we only sample the codes that trigger a jump in the sigmoid values for the various classes. Future work could focus on applying model explainability techniques (XAI) to enable analysis of attributing preceding code sequences when such jumps occur (e.g., using techniques like SHAP \citep{lundberg2017unified} or Integrated Gradients \citep{sundararajan2017axiomatic}).

From simply looking at changes to individuals' neighborhood in the latent space, in Figures \ref{fig:tragic_events}, we can see that a large change seems to correlate with changes to their overall health profile.
It is no surprise to see that childbirth causes large changes due to all the related health events. 
However, the tragic events in the target group also seem to have an impact on their health profiles, even though these are less obvious to see from such aggregated data. From analyzing some of the target individuals in more detail, we observed that several showed symptoms of depression and similar issues related to these events, but the relative onset time of these symptoms varied greatly.

In the experiment in Section~\ref{subsec:exp_nearest_neighbours_analysis} we demonstrate an alternative way of visualizing how a person's health profile evolves over time in terms of their similarity to various reference disease profiles in the latent embedding space.
These similarity plots, with their decreasing similarities over time, may also somewhat reflect the uniqueness of individuals' health trajectories, and that there is an overall increase in diversity or variability in the health characteristics of individuals as they age, as reported in \citet{jylhava2017biological}.
Associated with this, using another transformer-based approach, \citet{munoz2023interpreting} were able to identify and visualize multiple progression patterns among individuals with the same disease.
Going back to the present study, this visualization approach may help in monitoring and interpreting health trajectories. 
The underlying age-wise similarity matching in the embedding space also allows retrieval of similar reference individuals to a target person, which can, e.g., be helpful to clinicians in deciding future healthcare interventions.
This approach provides a different view of the health trajectories compared to looking at the changing sigmoid values, as in Figure~\ref{fig:sigmoid_and_nn_changes}.
Further research is required to determine whether this visualization approach has some additional or supplementary properties.
See Figures \ref{fig:PersonC} and \ref{fig:PersonD} in the Appendix for other examples of health trajectories visualized by sigmoid values and by age-wise embedding similarities to reference individuals.

Without going into much detail here, through some basic experimentation with class weighting during model training we observed that this would help make similarities and dissimilarities more prevalent (see Figure~\ref{fig:nearness_trajectory_ageWise_weighted} in the Appendix for an example).


Given that we condition the model with $\textrm{t2f}$ –– the time until the forecast interval for when to predict the diagnoses -- and given the diverse training data, the trained model should allow exploration of health trajectories relative to other forecast intervals simply by changing $\textrm{t2f}$ at inference time.
As another future work, it would be interesting to explore whether the model could aid in hypothesis generation, detect underdiagnosed diseases, or identify patterns that signal overdiagnosis. 
Another direction could be to explore ways to encourage the model to form more discriminative and disease-specific latent representations, e.g., by applying loss functions similar to those introduced in \citet{lee2020temporal}.

\paragraph{Limitations}
Due to resource and time limitations, we only performed basic hyperparameter optimization for the transformer-based models. It is possible that better performance could be gained, especially with a larger model. Further, we do not focus on class weight optimization. 
Finally, pre-training of transformer-based models has been widely successful, especially in natural language processing. We do not explore this here, mainly so because all of our data can be said to be labeled training data.

\section*{Acknowledgments}
We are grateful to Finnish individuals, whose data made this study possible. We would also like to thank the entire FinRegistry team for making the data available for the study. 
This study has received funding from the European Research Council (H2020 grant 101016775, 945733, and NextGenerationEU), and from the Research Council of Finland (Flagship programme: Finnish Center for Artificial Intelligence FCAI, and grants 352986, 358246, 323116).

\section*{Ethics approval}
FinRegistry is a collaboration project of the Finnish Institute for Health and Welfare (THL) and the Data Science Genetic Epidemiology research group at the Institute for Molecular Medicine Finland (FIMM), University of Helsinki. The FinRegistry project has received approvals for data access from THL (THL/1776/6.02.00/2019 and subsequent amendments), Digital and Population Data Services Agency (VRK/5722/2019–2), Finnish Center for Pension (ETK/SUTI 22003) and Statistics Finland (TK-53–1451-19). The FinRegistry project has received IRB approval from THL (Kokous 7/2019).

\bibliography{references}

\begin{appendices}

\section{Model training details}
\label{apd:model_training}
For training we used a server with a single NVIDIA A100 GPU with 40GB VRam. 
As hyper parameters for the transformer-based models we used a hidden layer (embedding) size of 384, 8 attention heads and 8 layers, with a maximum sequence length of 400 codes. 
Due to limited computing resources and time, we optimized the remaining hyper parameters with a basic grid search for learning rate, dropout, and batch size. 
For these models we ended up with the same hyper parameters, with a learning rate of $1e-3$ (we also used learning rate decay), a dropout of 0.1, and a batch size of 160 (which is the largest batch size that fits into VRam memory).
We use learnable embedding representations for age, position, and years to forecast, which are reported to work well for time-series data \citet{wen2023transformers}.
Given that the prevalence (class distribution) is highly imbalanced (see Prevalence column in Table~\ref{tab:prevalence}, Appendix~\ref{apd:second}), we used random downsampling of \emph{none} entries during training.
We also briefly explored applying some class weighting to the loss function. This did not seem to provide a positive impact on the prediction scores (at least not micro scores). However, we noticed that class weighting would make the latent embedding space more discriminating. Also, to give a fair comparison between the methods/models, we did not apply any class weighting when training the models used in Section~\ref{sec:experiments}.
We also used early stopping with a buffer of 10 epochs during training.
We acknowledge that there is room for more hyperparameter exploration, but we want to emphasize that the main focus of this work is not on predictive performances.

For the \texttt{XGBoost} and \texttt{LogReg} models, we optimized the hyperparameters through grid search.
\texttt{XGBoost} parameters were: learning rate = 0.1, max depth = 7, subsample = 1.0, colsample bytree = 0.5, and $n$ estimators = 150.
\texttt{LogReg} parameters were: penalty = l1, C = 0.1, and solver = liblinear.

\section{Disease prevalence}
\label{apd:second}

Table~\ref{tab:prevalence} shows prevalence for the different classes (diseases, death and none), together with their AUROC and Recall@4 scores, calculated with the \texttt{Evolve} model.

\begin{table*}[htb]
    \scriptsize 
    \centering
    \caption{Table shows the individual AUROC and Recall@4 scores for each class, as well as their prevalence. These scores come from the \texttt{Evolve} model.}
    \begin{tabular}{llll|llll}
        \toprule
        \textbf{Class} & \textbf{AUROC} & \textbf{Recall@4} & \textbf{Prevalence} & \textbf{Class} & \textbf{AUROC} & \textbf{Recall@4} & \textbf{Prevalence}  \\
        \midrule
        None & 0.8012 & 0.9944 & 0.8506 & Gout & 0.9145 & 0.4292 & 0.0042\\
        Death & 0.9386 & 0.7497 & 0.0433  & Breast cancer & 0.8616 & 0.3136 & 0.0037\\
        Arterial fibrillation & 0.8987 & 0.8029 & 0.0223 & Prostate cancer & 0.9331 & 0.0962 & 0.0036\\
        Cancer (non-specific) & 0.8164 & 0.7195 & 0.0199 & Epilepsy & 0.7495 & 0.1739 & 0.0034\\
        Type 2 diabetes & 0.8494 & 0.7329 & 0.0189 & Colorectal cancer & 0.8256 & 0.0009 & 0.0021\\
        Depression & 0.8232 & 0.8292 & 0.0171 & Lung cancer & 0.8961 & 0.0035 & 0.0020\\
        Coronary hearth disease & 0.8678 & 0.5900 & 0.0161 & Seropositive RA & 0.7705 & 0.1381 & 0.0015\\
        Knee arthrosis & 0.8643 & 0.3905 & 0.0118 & Interstitial lung disease & 0.8591 & 0.0007 & 0.0014\\
        Asthma & 0.8435 & 0.7221 & 0.0107 & Melanoma skin cancer & 0.7234 & 0.0000 & 0.0011\\
        Coxarthrosis & 0.8456 & 0.2560 & 0.0080 & Type 1 diabetes & 0.8067 & 0.2626 & 0.0006\\
        Appendicitis & 0.6420 & 0.5500 & 0.0055 &  &  &  & \\
        \bottomrule
    \end{tabular}
    \label{tab:prevalence}
\end{table*}

\section{Trajectories and predictive codes}
\label{apd:third}

\begin{figure*}[htb]
      \centering
      \includegraphics[width=0.9\linewidth]{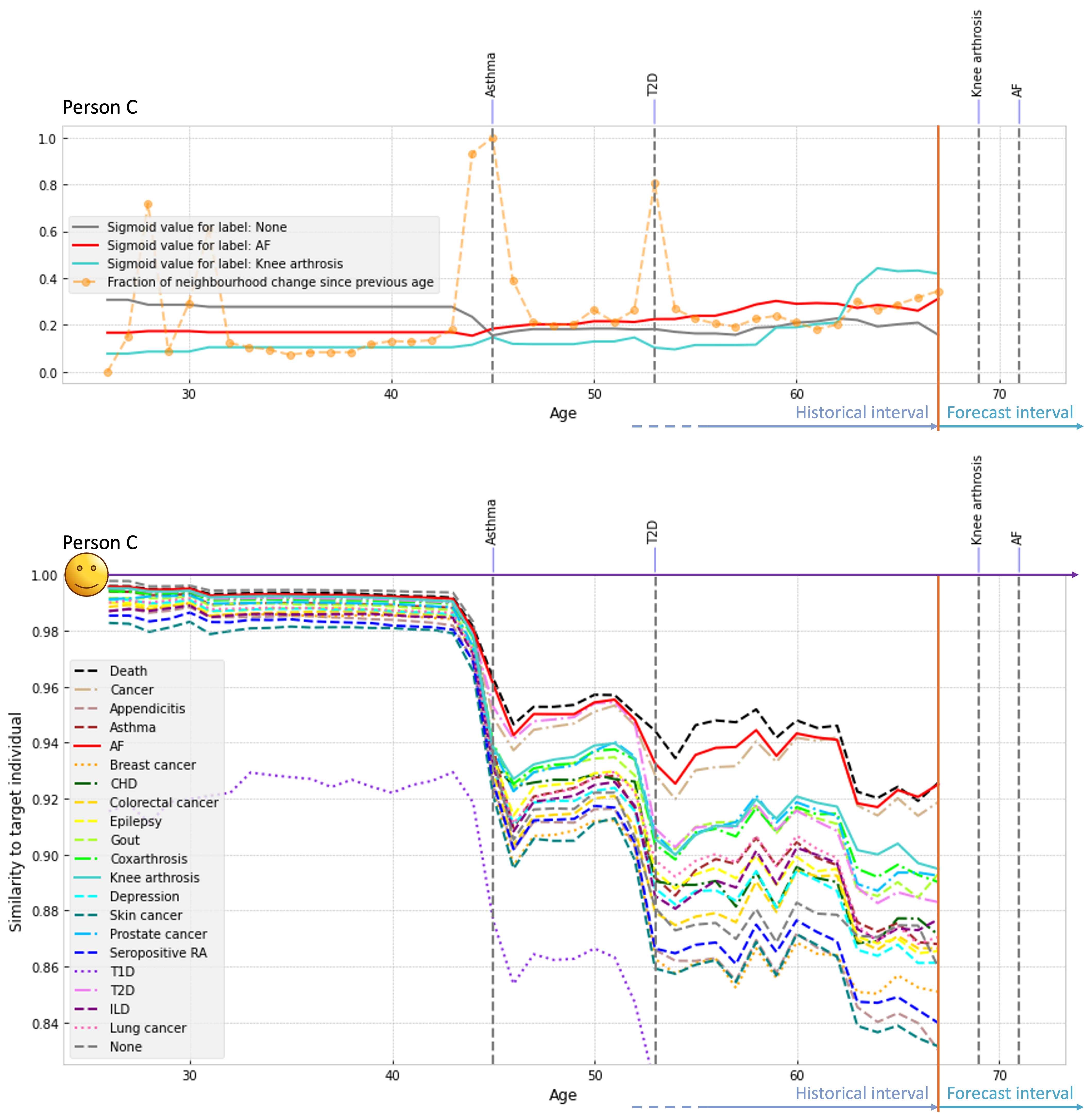}
      \caption{The figure shows the evolving health trajectory of a person who will end up with knee arthritis and atrial fibrillation in the given forecast interval. Top plot shows the changing sigmoid activations through the health trajectory, while the bottom plot shows age-wise similarities between the target individual and the top $k=100$ most similar representatives from each diagnosis (class).}
      \label{fig:PersonC}
\end{figure*}

\begin{figure*}[htb]
      \centering
      \includegraphics[width=0.9\linewidth]{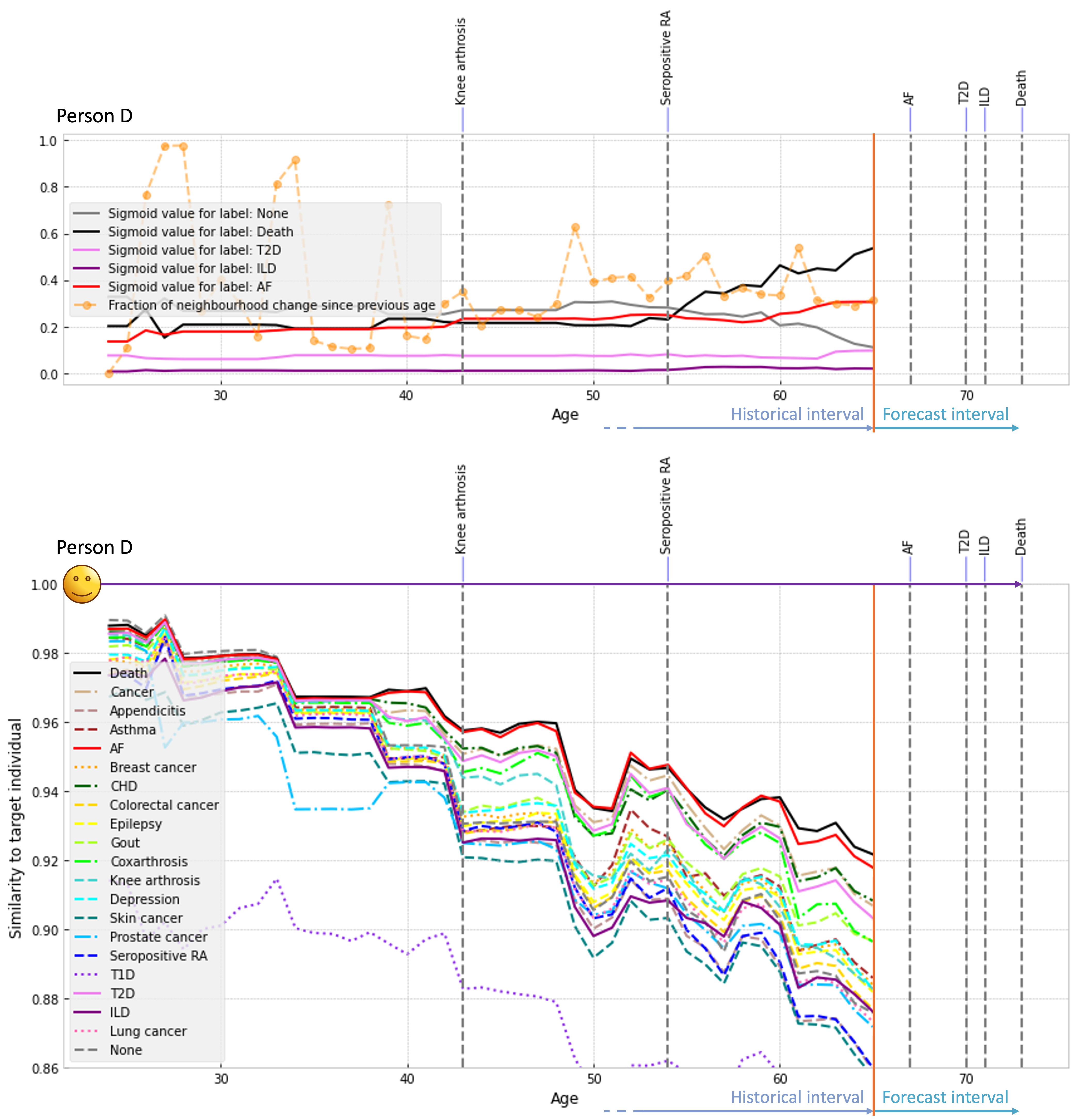}
      \caption{The figure shows the evolving health trajectory of a person who will get multiple diseases and die in the forecast interval. Top plot shows the changing sigmoid activations through the health trajectory, while the bottom plot shows age-wise similarities between the target individual and the top $k=100$ most similar representatives from each diagnosis (class).}
      \label{fig:PersonD}
\end{figure*}

Figures \ref{fig:PersonC} and \ref{fig:PersonD} shows examples of health trajectories visualized by sigmoid values and by age-wise embedding similarities to reference individuals.

\begin{figure*}[htb]
      \centering
      \includegraphics[width=0.9\linewidth]{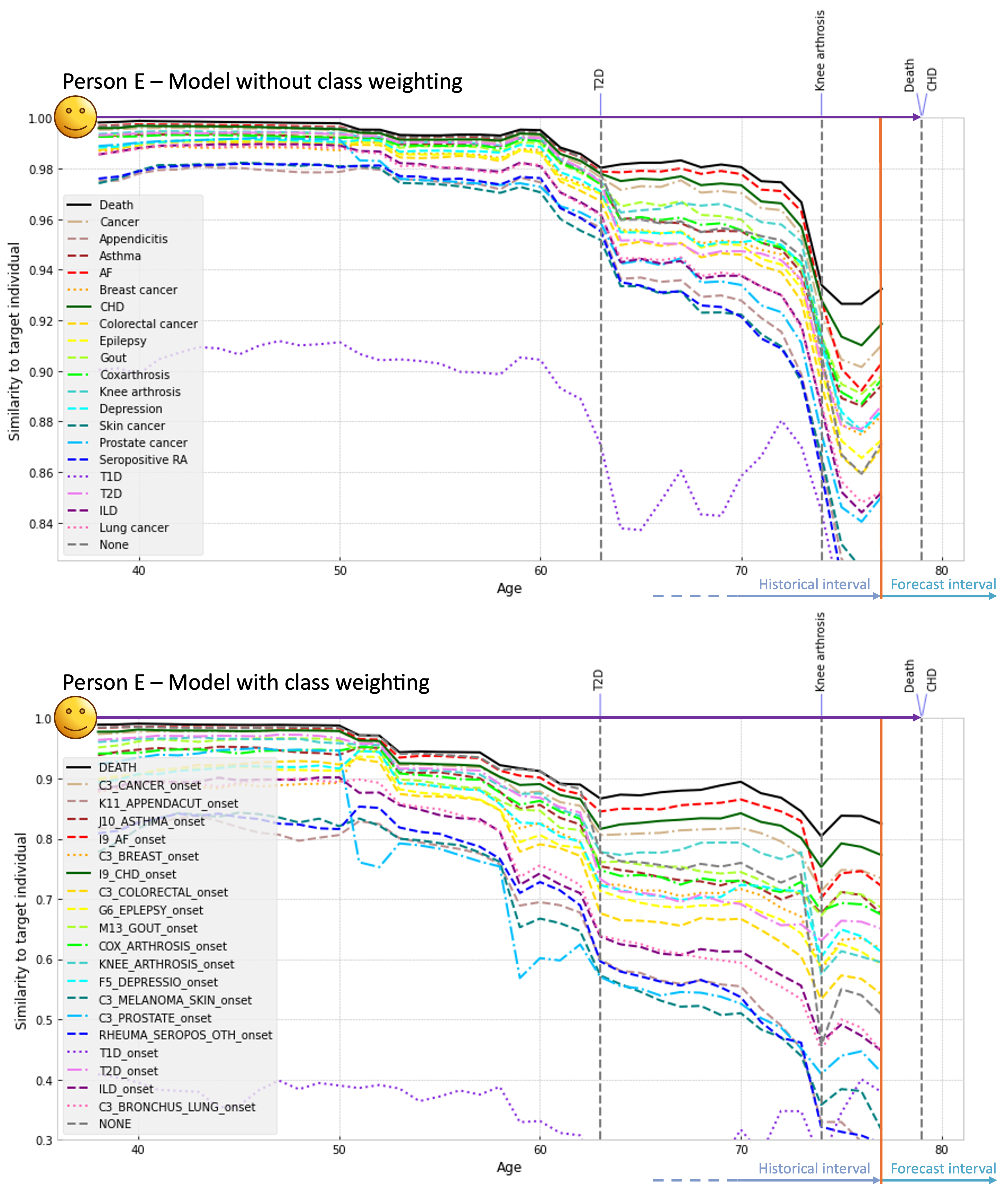}
      \caption{This illustrates the differences between not using class weighting (top) and using class weighting (bottom) when training the \texttt{Evolve} model. As can be seen from the Y-axis, class weighting results in an embedding space that is more discriminative and with more prevalent similarities and dissimilarities.}
      \label{fig:nearness_trajectory_ageWise_weighted}
\end{figure*}

Figure~\ref{fig:nearness_trajectory_ageWise_weighted} shows an example of how applying class weighting during model training (bottom plot) results in an embedding space that is more discriminative and with more prevalent similarities and dissimilarities.


Table~\ref{tab:early_examples_cont} shows the most frequent diagnosis and drug codes that triggers jumps in the sigmoid values for the classes not shown in Table~\ref{tab:early_examples}.
\begin{table*}[htb]
    \centering
    \scriptsize 
    \caption{Most frequent diagnosis codes and drug codes that triggers jumps in the sigmoid values of the respective classes.}
    \begin{tabular}{l|llll|llll}
        \toprule
        \textbf{Class} & \textbf{Diagnosis} & \textbf{\%} & \textoverline{\textbf{age}} & \textoverline{\textbf{t2f}} & \textbf{Drug} & \textbf{\%} & \textoverline{\textbf{age}} & \textoverline{\textbf{t2f}}  \\
        \midrule
        & Alzheimer’s disease & 5.23 & 76 & 5                               & Dopa and dopa derivatives & 1.63 & 72 & 6 \\
        Death & Malignant neoplasm & 4.44 & 68 & 3                          & Anticholinesterases & 1.44 & 76 & 5 \\
        & Dementia & 3.72 & 76 & 5                                          & Pyrimidine analogues & 0.94 & 63 & 3 \\
        \midrule
        & Type 2 diabetes & 5.60 & 66 & 5                                           & Beta blocking agents, selective & 3.07 & 72 & 5 \\
        Cancer (non specific) & Atrial fibrillation and flutter & 1.81 & 66 & 10    & HMG CoA reductase inhibitors & 2.93 & 73 & 4 \\
        & Depression & 0.82 & 57 & 7                                                & Beta blocking agents, selective & 2.72 & 75 & 5 \\
        \midrule
        & Depression & 2.62 & 26 & 7                                        & Propionic acid derivatives & 1.62 & 29 & 4 \\
        Appendicitis & Asthma & 1.36 & 37 & 6                               & First-generation cephalosporins & 1.24 & 26 & 4 \\
        & Malignant neoplasm & 0.86 & 38 & 7                                & Insulins and analogues for injection & 1.22 & 35 & 11 \\
        \midrule
        & COPD & 8.55 & 51 & 16                                             & Glucocorticoids & 26.66 & 58 & 3 \\
        Asthma & Childhood allergy & 0.48 & 3 & 1                           & Selective beta-2-adrenoreceptor ... & 26.37 & 55 & 4 \\
        & Type 2 diabetes & 0.45 & 69 & 4                                   & Adrenergics + corticosteroids ... & 13.69 & 61 & 4 \\
        \midrule
        & M. coronary heart disease event & 3.10 & 67 & 9                & Beta blocking agents, selective & 5.69 & 72 & 4 \\
        Atrial fibrillation & Heart failure, not strict & 2.28 & 68 & 5     & Vitamin K antagonists & 4.76 & 76 & 3 \\
        & Type 2 diabetes & 1.59 & 69 & 5                                   & HMG CoA reductase inhibitors & 4.13 & 77 & 4 \\
        \midrule
        & Single spontaneous delivery & 6.57 & 28 & 30                      & Progestogens and estrogens & 31.71 & 59 & 7 \\
        Breast cancer & Noninflammatory disorders of & 3.66 & 47 & 19       & Natural and semisynthetic estrogens & 5.46 & 60 & 10 \\
        & female genital tract & & &                                        & & & & \\   
        & Pregnancy with abortive outcome & 4.25 & 28 & 31                  & Intrauterine contraceptives & 1.15 & 50 & 4 \\
        \midrule
        & Type 2 diabetes & 5.39 & 62 & 7                                   & Beta blocking agents, selective & 5.53 & 73 & 6 \\
        Colorectal cancer & Asthma & 3.06 & 63 & 14                         & HMG CoA reductase inhibitors & 5.31 & 74 & 5 \\
        & Coxarthrosis & 0.98 & 70 & 8                                      & Glucocorticoids & 4.35 & 71 & 9 \\
        \midrule
        & Heart failure, not strict & 1.41 & 69 & 4                         & Uric acid inhibitors & 55.33 & 63 & 9 \\
        Gout & Type 2 diabetes & 1.33 & 65 & 5                              & Beta blocking agents, selective & 1.14 & 75 & 2 \\
        & Gonarthrosis & 0.75 & 68 & 6                                      & HMG CoA reductase inhibitors & 0.79 & 74 & 2 \\
        \midrule
        & Gonarthrosis & 10.71 & 53 & 11                                     & Coxibs & 4.95 & 66 & 2 \\
        Coxarthrosis & Asthma & 1.82 & 63 & 6                               & Anti-inflammatory [...] agents & 3.28 & 56 & 8 \\
        & Varicose veins & 1.44 & 39 & 29                                   & Anilides & 1.92 & 70 & 2 \\
        \midrule
        & Disorders of the thyroid gland & 2.08 & 70 & 1                    & Anti-inflammatory [...] agents & 7.99 & 57 & 8 \\
        Knee arthrosis & Asthma & 1.58 & 59 & 5                             & Anilides & 2.56 & 68 & 1 \\
        & Internal derangement of knee & 0.59 & 50 & 6                      & Propionic acid derivatives & 2.21 & 62 & 2 \\
        \midrule
        & Malignant neoplasm & 1.46 & 71 & 5                                & Beta blocking agents, selective & 4.44 & 71 & 8  \\
        Skin cancer & Malignant neoplasm of prostate & 0.73 & 68 & 6        & HMG CoA reductase inhibitors & 4.44 & 71 & 4 \\
        & Hypertension & 0.70 & 63 & 14                                     & Uric acid inhibitors & 2.90 & 66 & 10 \\
        \midrule
        & Type 2 diabetes & 3.89 & 70 & 4                                   & Alpha-adrenoreceptor antagonists & 9.73 & 72 & 3 \\
        Prostate cancer & Hyperplasia of prostate & 1.91 & 69 & 4           & Testosterone-5-alpha r. inhibitors & 8.99 & 70 & 5 \\
        & Diseases of male genital organs & 1.53 & 65 & 11                  & Beta blocking agents, selective & 2.01 & 73 & 4 \\
        \midrule
        & Polyarthropathies & 2.36 & 58 & 6                                     & Aminoquinolines & 6.63 & 58 & 6  \\
        Seropositive RA & Seronegative rheumatoid arthritis & 1.61 & 62 & 7     & Natural and semisynthetic estrogens & 2.87 & 65 & 10 \\
        & Arthropathies & 1.43 & 59 & 7                                         & Beta blocking agents, selective & 2.44 & 68 & 6 \\
        \midrule
        & Asthma & 3.21 & 4 & 4                                             & Insulins and analogues for injection* & 12.26 & 56 & 9  \\
        Type I diabetes & Chronic pancreatitis & 1.92 & 40 & 8              & HMG CoA reductase inhibitors & 1.21 & 60 & 7 \\
        & Acute pancreatitis & 1.31 & 38 & 8                                & ACE inhibitors, plain & 1.05 & 50 & 9 \\
        \midrule
        & Gestational diabetes (for exl.) & 12.54 & 32 & 12                 & Uric acid inhibitors & 3.82 & 48 & 7  \\
        Type II diabetes & Obesity & 3.23 & 45 & 5                          & Beta blocking agents, selective & 1.95 & 60 & 3 \\
        & Acute pancreatitis & 1.43 & 49 & 8                                & HMG CoA reductase inhibitors & 1.91 & 61 & 3 \\
        \midrule
        & Seropositive RA & 6.72 & 64 & 9                                   & Beta blocking agents, selective & 3.00 & 72 & 5 \\
        ILD & Polyarthropathies & 3.63 & 65 & 8                             & Aminoquinolines & 2.58 & 66 & 8 \\
        & Arthropathies & 1.40 & 66 & 8                                     & HMG CoA reductase inhibitors & 2.37 & 72 & 4 \\
        \midrule
        & COPD & 4.74 & 67 & 8                                              & Anticholinergics & 5.80 & 68 & 7  \\
        Lung cancer & Atherosclerosis ... & 3.06 & 64 & 10                  & Adrenergics + corticosteroids ... & 3.29 & 68 & 6 \\
        & Type 2 diabetes & 1.20 & 67 & 5                                   & Selective beta-2-adrenoreceptor ... & 2.52 & 67 & 6 \\
        \midrule
        & Type 2 diabetes & 4.35 & 56 & 4                                   & Natural and semisynthetic estrogens & 2.24 & 58 & 14  \\
        None & Depression & 4.17 & 28 & 7                                   & HMG CoA reductase inhibitors & 0.82 & 62 & 5 \\
        & Single spontaneous delivery & 3.03 & 30 & 29                      & Progestogens and estrogens & 0.80 & 53 & 19 \\
        \bottomrule
    \end{tabular}
    \label{tab:early_examples_cont}
\end{table*}

\end{appendices}

\end{document}